\documentclass{article}
\usepackage{amsmath,amssymb}
\usepackage{multirow}
\usepackage{graphicx}
\usepackage{algorithm}
\usepackage{algorithmic}
\usepackage{color,soul}
\usepackage{booktabs} % For formal tables
\usepackage{tabularx}
\usepackage{subcaption}
\usepackage{mwe}
\usepackage{commath}
\usepackage{verbatim}
\usepackage{mathtools}
\usepackage{cuted}
\usepackage{flushend}
\usepackage{arydshln}
\usepackage{multirow}

\usepackage{cite}
\usepackage{amsmath,amssymb,amsfonts}
\usepackage{algorithmic}

\usepackage{textcomp}
\usepackage{xcolor}
\usepackage{authblk}
\usepackage{times}  %Required
\usepackage{helvet}  %Required
\usepackage{courier}  %Required
\usepackage{url}  %Required
\usepackage{graphicx}  %Required

\usepackage{mathtools}
\usepackage{amsmath,amssymb,amsfonts}

\usepackage{caption}
\usepackage{amsthm,dsfont}
\newcommand{\ignore}[1]{}
\usepackage{color}
\usepackage{authblk}
\usepackage{graphicx}
\usepackage{float} 
\usepackage{multirow}
\usepackage{amsfonts}
\usepackage{comment}
\usepackage{fancyhdr}
\usepackage{rotating}

\def\BibTeX{{\rm B\kern-.05em{\sc i\kern-.025em b}\kern-.08em
    T\kern-.1667em\lower.7ex\hbox{E}\kern-.125emX}}
\begin{document}

\title{On the Behaviour of Differential Evolution for Problems with Dynamic Linear Constraints}

\author{Maryam Hasani-Shoreh}
\author{Mar\'{\i}a-Yaneli Ameca-Alducin}
\author{Wilson Blaikie}
\author{Frank Neumann}
%\author{Marc Schoenauer}
\affil{Optimisation and Logistics, School of Computer Science,\\ The University of Adelaide, Adelaide, Australia}

\author{Marc Schoenauer}
\affil{INRIA, LRI, CNRS, University of Paris-Saclay, France }

\renewcommand\Authands{ and }
\maketitle
\begin{abstract}
Evolutionary algorithms have been widely applied for solving dynamic constrained optimization problems (DCOPs) as a common area of research in evolutionary optimization. Current benchmarks proposed for testing these problems in the continuous spaces are either not scalable in problem dimension or the settings for the environmental changes are not flexible. Moreover, they mainly focus on non-linear environmental changes on the objective function. While the dynamism in some real-world problems exists in the constraints and can be emulated with linear constraint changes. The purpose of this paper is to introduce a framework which produces benchmarks in which a dynamic environment is created with simple changes in linear constraints (rotation and translation of constraint's hyperplane). 
Our proposed framework creates dynamic benchmarks that are flexible in terms of number of changes, dimension of the problem and can be applied to test any objective function. 
Different constraint handling techniques will then be used to compare with our benchmark. 
The results reveal that with these changes set, there was an observable effect on the performance of the constraint handling techniques.
\end{abstract}

\section{Introduction}
\label{sec:Intro}
Dynamic constrained optimization problems (DCOPs) in which the objective function or/and the constraints change over time is observed in a variety of real world problems. Examples include hydro-thermal power scheduling~\cite{de2008plant}, source identification~\cite{liu2008adaptive}, and parameter estimation~\cite{prata2006simultaneous} in which the dynamism arise because the available resources or demand vary over time, the information about the problem is gradually revealed, or parameter tuning is needed as time passes.

Multiple evolutionary algorithms have been designed so far to solve these problems~\cite{ECDCOPs, Nguyen20121,Yaneli2016}. The focus of these papers are either on dealing with dynamism in the environment including introducing~\cite{Goh_2009} or maintaining diversity~\cite{Bui2005}, memory-based approaches~\cite{Richter2013} and multi-population approaches~\cite{branke2000multi}, or mechanisms to deal with constraints including penalty~\cite{CEC09}, repair methods~\cite{Das,bu2017continuous,nguyen2012continuous} and feasibility rules~\cite{Yaneli2016}.
In addition, some papers enhanced both constraint handling and dynamic handling mechanisms~\cite{Nguyen20121}.
Among the many evolutionary algorithms, DE has showed competitive results in dynamic and constrained optimization problems so far~\cite{ameca2018comparison}.

However, besides to developing algorithms there should be a comprehensive benchmark suit that can test algorithms considering a range of characteristics.
Although there are a range of benchmarks proposed to test the relevant algorithms for discrete spaces~\cite{roostapour2018performance}, and/or multi-objective optimization in dynamic environments~\cite{jiang2017evolutionary}, for continuous spaces in single objective optimization so far, the most used benchmark is the proposed benchmark in~\cite{Nguyen20121}. In this benchmark, the dynamic changes are applied by adding time-dependent terms to the objective function and the constraints of one of the functions ($G\_24$) of the static benchmark proposed in CEC 2006~\cite{liang2006problem}. However, there are parameters defined to alter the severity of changes in the environment, this benchmark is based on one objective function and the transformation of this function and is not applicable to test different functions to consider a range of characteristics. Moreover, the proposed problem is a two-dimensional in size and is not flexible to be applied for larger dimension of the problem.
In addition, the feasible regions of the dynamic constraint function in this benchmark are very large, which might not be sufficiently complicated. Bu et all in~\cite{bu2017continuous}, introduces one variant of this benchmark suit that have a parameter to control the size and the number of the feasible regions. The other variant introduced in this paper is based on the moving peak benchmark.

A similar benchmark is proposed in~\cite{zhang2014danger} that is based on dynamic transformations introduced by Nguyen in~\cite{Nguyen20121}. However, the problem information, including the number of feasible regions, the global optimum, and the dynamics of each feasible region, is lacking. The lack of such information makes it difficult to measure and analyze the performance of an algorithm and probably this is the reason this benchmark have become less popular than Nguyen benchmark~\cite{Nguyen20121}.

In terms of having a scalable and flexible benchmark, in the literature there are some benchmark generators proposed. Like in~\cite{li2008benchmark} a dynamic benchmark generator is proposed that is designed with the idea of constructing dynamic environments across binary, real, and combinatorial solution spaces. The dynamism is obtained by tuning some system control parameters, creating six change types: small step, large step, random, chaotic, recurrent, and recurrent change with noise. 
 
While the aforementioned benchmark generator's main focus is on creating dynamic objective functions, in this paper we put our focus on creating dynamism in the constraints. Our motivation comes from characteristic of some real-world problems like scheduling power system problem having dynamic linear constraints (due to the variable demand and available resources over-time). For a better insight about the effects of constraint changes we keep the objective function static. Indeed, this is the case in some real world problems in which only constraints will change like the problem of hydro-thermal power scheduling in continuous spaces~\cite{deb2007dynamic} or the ship scheduling problem in discrete spaces~\cite{mertens2006dyncoaa}.

Dynamic changes are imposed by the translation and rotation of the constraint's hyperplane.
The examples of these two operations on the constraint in a real-world dynamic environment are: the reduction and increment of demand that happens regularly at power system (hyperplane translation) or changes on the share of each plant power production (hyperplane rotation)~\cite{morales2014managing}.

Our proposed benchmark generator is flexible (frequency and severity of changes, number of environmental changes, and dimension of the problem), simple to implement (with any objective function), analyze, or evaluate and computationally efficient and finally allows conjectures to real-world problems.
 
In the experiments we apply differential evolution (DE) algorithm with different constraint handling techniques and observe how they deal with these changes depending on the magnitude and frequency of changes. 

Our experiments are repeated across some well-known functions including sphere, Rastrigin, Ackley and Rosenbrock.
For the analysis on the performance of the tested algorithms, a ranking procedure is introduced that uses the values of the objective function and the constraint violations to rank the performance of the algorithms. In addition, the common measure modified offline error is also evaluated for the experiments and the results are investigated.
The results reveal that the changes on frequency and hyperplane rotation and translation have a direct correlation with the performance of the constraint handling techniques. Therefore with imposing simple linear changes we could effectively put the algorithms to struggle and test their performance.

The outline of the paper is as follows. Section~\ref{sec:Prim} introduces a short overview of the problem statement. In Section~\ref{sec:change-setup}, our proposed dynamic changes' framework is described. Experimental investigations will be presented in Section~\ref{sec:Experiment} and finally in Section~\ref{sec:conclusion} conclusions and future work are summarized.

\section{Problem statement}
\label{sec:Prim}
In this section a general overview of the problem statement is presented.
\subsection{Dynamic constraint optimization problems}
A dynamic constrained optimization problem (DCOP) is an optimization problem where the objective function and/or the constraints can change over time~\cite{Nguyen20121}. Such an optimization problem ideally must be solved at every time instant $t$ or whenever there is a change in any of the objective function and/or the constraints with $t$. In such optimization problems, the time parameter can be mapped with the iteration counter $\tau$ of the optimization algorithm. Such problems often arise in real-world problem solving, particularly in optimal control problems or problems requiring an on-line optimization~\cite{de2008plant}.

Generally, the problem statement in DCOPs can be defined as follows. 

Find $\vec{x}$, at each time $t$, which: 
\begin{equation}
		\min_{\vec{x}\in F_t\subseteq[L, U]} f(\vec{x}, t)
\end{equation}

where $f:S \rightarrow \mathbb{R}$ is a single objective function, $\vec{x} \in \mathbb{R}^D $ is a solution vector and $t \in N^+$ is the current time, 
\begin{equation}
[L, U]=  \left\lbrace \vec{x} = (x_{1},x_{2},...,x_{D}) \mid L_i \leq x_i \leq U_i, i = 1 \ldots D\right\rbrace
\end{equation}
is called the search space ($S$), where $L_i$ and $U_i$ are the lower and upper boundaries of the $i$th variable,

subject to:
\begin{equation}
\begin{array}{l}
\label{eq:searchspace}
F_{t}=\{ \vec{x} \mid \vec{x} \in [L,U], g_i (\vec{x},t) \le 0, i = 1, \ldots, m,\\
h_j (\vec{x},t) = 0,j = 1, \ldots, p\} \\
\end{array}
\end{equation}
is called the feasible region at time $t$, where $g_i(x, t)$ is the linear $i$th inequality constraint at time $t$ and $h_j(x,t)$ is the $j$th equality constraint at time $t$.

$\forall \vec{x} \in F_t$, if there exists a solution $\vec{x}^* \in F_t$ such that 
$f(\vec{x}^*,t)\leq f(\vec{x},t)$,  then $\vec{x}^*$ is called a feasible optimal solution 
and $f(\vec{x}^*,t)$ is called the feasible optimal value at time $t$.

\section{Dynamic changes framework}
\label{sec:change-setup}
Many real-world problems lie in the area of DCOPs, examples include hydro-thermal scheduling problem~\cite{deb2007dynamic}, source identification~\cite{liu2008adaptive} and parameter estimation~\cite{prata2006simultaneous}. Many of these real-world problems have single or multiple linear constraints. Therefore, the relevant benchmark can be as simple as creating some changes in the constraints coefficients and boundaries that represent changes in different times. In this section we will introduce a framework to create changes on linear constraints to emulate a dynamic environment.
Our proposed changes is observed in some real-world problems like power scheduling problem~\cite{morales2014managing}, in which the conditions in the system like demand or available resources will change. In this section first the constraint setup is presented and then the frequency setup will be explained.
\subsubsection{Constraint setup}
For emulating the dynamic constraints, simple linear constraints are used for search space modification. Although linear constraints are simple, they used as a representation of some of the real world problem constraints and prevents over-complication of the analysis. The general formulation for the linear constraints are as follows. 
 \begin{equation}
g_i(\vec{x})=\sum\limits_{j=1}^{D} a_j x_j-b_i \le 0 \quad  i \in\{ 1, \ldots, m\} 
 \end{equation}
where $g_i(\vec{x})$ is the $i$th inequality constraint, $a_j$ is the $j$th variable coefficient, $x_j$ is the $j$th decision variable, $b_i$ is the upper limit of the $i$th  constraint and $D$ is equal to the problem dimension.
A general case for one constraint is defined first and then is developed for multiple constraints. 

Two operations are defined for changes on constraint, the first one is related to changes on $b$ coefficient (hyperplane translation) and the second one is changes of $a_i$ coefficients (hyperplane rotation). 
These two changes in the simple linear constraint can happen commonly in real-world problems. In a scheduling power plant problem, the changes of demand or capacity of each production plants including stochastic renewable plants can be an example of hyperplane translation. The changes on the share of each plant to produce overall supply, or the probability of renewable resources production can be an example of hyperplane rotation (these are the coefficient of variables). Our following proposal for creating changes in the environment is generally showing the usual changes in some real-world dynamic problems like power system problem. 
 
\textbf{Hyperplane translation:}
If $a_i$ coefficients is chosen in a way to create a unit normal vector of $\vec{a}$, changes of $b$ will directly show the effects of changing the distance from the optimum point\footnote{for all the chosen functions zero is the optimum point}. The distance $d$ of the constraint hyperplane from the origin $0^D$ is given by
$	d=b/\left(\sum\limits_{i=1}^{D} a_i^{2}\right)^{1/2}.$

\begin{figure}[t]
        \centering          \includegraphics[width=0.5\textwidth, height=4cm]{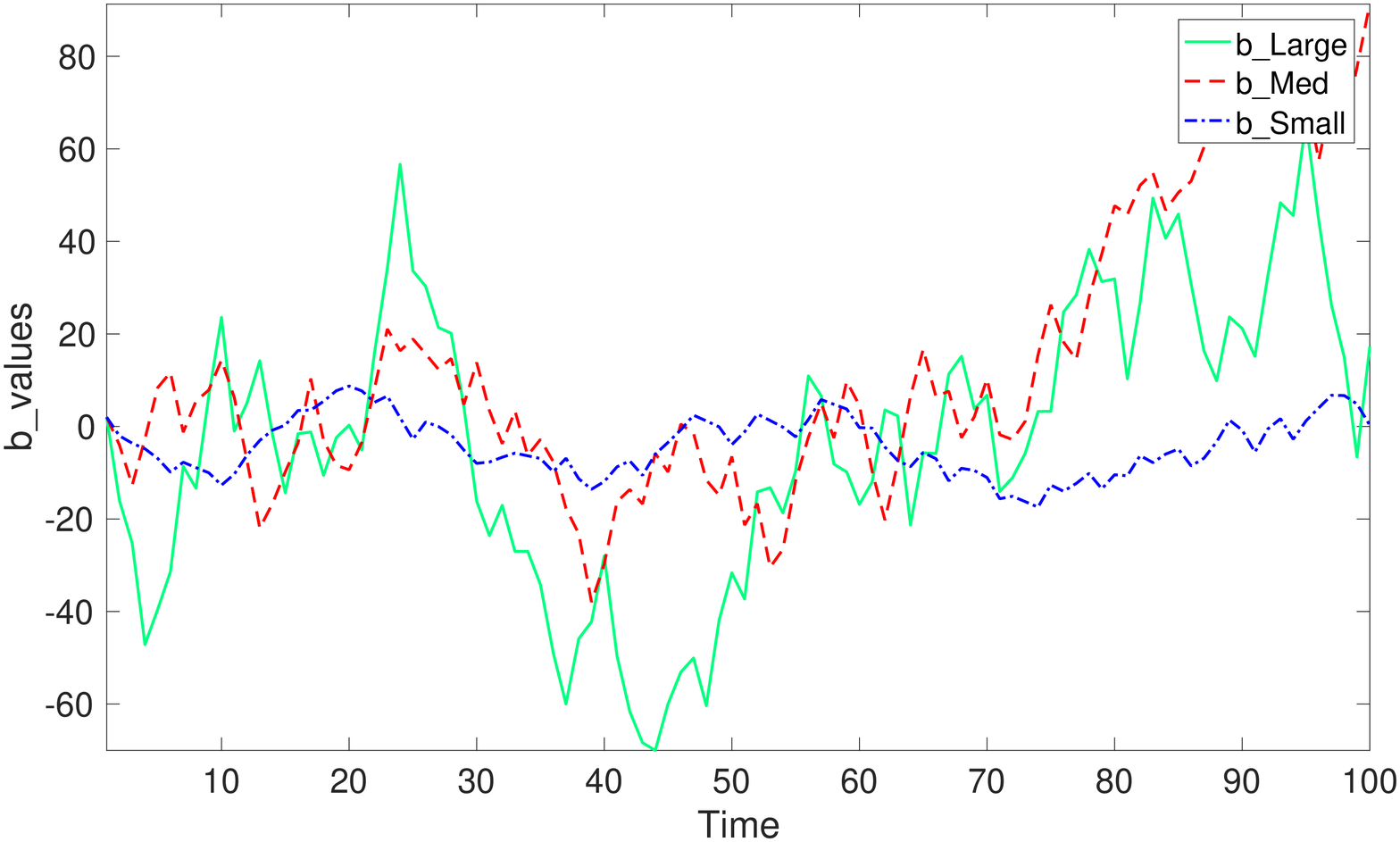}  
        \caption[]
        {\scriptsize  Sample settings for large, medium and small changes on $b$\_values} 
        \label{fig:changesofb}
    \end{figure}

The constraint bound value ($b(t)$) at time $t$ is obtained by adding a random value to its previous time value ($b(t-1)$) as in Equation~\ref{eq:bvalue}.  
\begin{equation}
\label{eq:bvalue}
	b(t)=b(t-1)+k_{r}
\end{equation}

where 
$k_{r}$
%=rand[0,1](h-l)+l$ 
is chosen uniformly at random within the interval $[lk,uk]$. Figure~\ref{fig:changesofb} shows an example of these settings for creating different magnitudes of change for single constraint case. 
These values are only samples of constraints boundaries for creating dynamic environment. As mentioned before, we can create multiple benchmarks for testing the algorithms with different scales of the changes based on the problem type. 
The two criteria that will affect the proper choice of the values of $lk$ and $uk$ are the dimension of the problem and the variable ranges.

\textbf{Hyperplane rotation:}
In this operation, the changes are created by rotating the hyperplane at each separate time. Random numbers are created in $\in [0,1]$ such that we have a unit normal vector of $\vec{a_i}$.
For any time we randomly select some of the coefficients and swap their values. As in this way $\vec{a}$ is still a unit normal vector, therefore the changes at each time is only related to the rotation and not the translation of hyperplane. 
So by making these changes at each time we will have a rotated hyperplane and we can observe how it will effect the algorithms behaviour. 

In another setting both changes of $a_i$ and $b$ values are considered. In this way in any time we have changes on either $b$ or $a_i$ based on a known probability.

With the current settings of the changes on the linear constraint coefficient, we can observe how the algorithms response to the new modified search space. With the comparisons we will observe which one of the compared algorithms will react faster in order to converge to the optimum after a change occurs. 
The coefficients of the linear constraints are generated through a proposed constraint generator and are then fed to the algorithm. Since the constraint coefficients are random, for a fair algorithms comparison, it is needed that all the generated coefficients for each run be the same for all the algorithms and a per algorithm generation would lead to a difference in compared constraints.
At each environment change, the number of feasible and infeasible solutions will change so that we expect to observe how efficiently the algorithms will manage the new set of constraints within the problem.

In many real-world problems including hydro-thermal scheduling problem, multiple constraints rather than single constraint will define the dynamic environment of the problem. In this case changes are imposed on $b_i$ value of $i$th constraint. In order to avoid complexity, at each time we only change one of the constraints boundaries, and this is aligned with the real-problems that one or a few criteria and not all will change at the new environment condition.   
\subsubsection{Frequency setup}
\label{subsec:FrequencySetup}

The frequency of change ($\tau$) is defined as how often the problem changes. A higher frequency seems to be more difficult for an algorithm to solve the related problem as less time is available at each period to reach the new global optimum~\cite{rohlfshagen2009dynamic}.
In the literature of DCOPs, the number of fitness evaluations is considered as a criteria to represent how frequently a change occurs~\cite{nguyen2012continuous}. 

\section{Differential evolution algorithm for dynamic constrained optimization}
In this section differential evolution algorithm, the applied constraint handling techniques and change detection mechanism are briefly introduced.

Differential evolution (DE) is a stochastic search algorithm that is simple, reliable and fast and showed competitive results in constraint and dynamic optimization~\cite{ameca2018comparison}. Each vector $\vec{x}_{i, G}$ in the current population (called at the moment of the reproduction as target vector) generates one trial vector $\vec{u}_{i, G}$ by using a mutant vector $\vec{v}_{i,G}$. The mutant vector is created applying $\vec{v}_{i,G}= \vec{x}_{r0,G} + F (\vec{x}_{r1,G} - \vec{x}_{r2,G})$,
where $\vec{x}_{r0,G}$, $\vec{x}_{r1,G}$, and $\vec{x}_{r2,G}$ are vectors chosen at random from the current population ($r0 \neq r1 \neq r2 \neq i$); $\vec{x}_{r0,G}$ is known as the base vector and $\vec{x}_{r1,G}$, and $\vec{x}_{r2,G}$ are the difference vectors and $F>0$ is a parameter called scale factor. Then the trial vector is created by the recombination of the target vector and mutant vector using a crossover probability $CR \in [0,1]$. 

In this paper a simple version of DE called DE/rand/1/bin variant is chosen~\cite{Mezura10a}; where ``rand" indicates how the base vector is chosen (at random in our case), ``1" represents  how many vector pairs will contribute in differential mutation, and ``bin" is the type of crossover (binomial in our case). 
Three different constraint handling techniques including penalty~\cite{tessema2009adaptive}, feasibility rules~\cite{deb2000efficient} and $\epsilon$-constrained~\cite{takahama2005constrained} are chosen to be included as for handling constraint with DE algorithm. With different constraint handling techniques we will observe how these algorithms will respond to the new changes in the environment.

The way that penalty functions treat with the infeasible solution and the amount of their strictness is largely dependent on the penalization factor. In this paper we used an adaptive penalty function method in which information gathered from the search process will be used to control the amount of penalties added to infeasible individuals~\cite{tessema2009adaptive}. In addition this adaptive penalty method uses the normalized values of objective function as we have different scales of objective values for different tested functions.
Applied Feasibility method is based on these three rules: i) between two feasible solutions, the one with the highest fitness 
value is selected, ii) if one solution is feasible and the other one is infeasible, the 
feasible solution is selected, iii) if both solutions are infeasible, the one with the lower sum of constraint violation is selected~\cite{deb2000efficient}.

The $\epsilon$-constrained method is similar to the feasibility rules with a more progressive manner toward dealing with constraints~\cite{takahama2005constrained}. In this method the solutions will be compared based on their objective values for those that have constraint violation below an epsilon level. The applied $\epsilon$-level is adaptive in such a way that for the first generations has larger values and gradually decreases to avoid infeasible solutions at the final generations. In order to adapt the $\epsilon$-level in dynamic optimization, the value of $\epsilon$ is re-initialized when the algorithms detect the change. In addition to the above-mentioned methods that relate to the constraint handling part of the algorithms, they need to be equipped with a mechanism to detect the environment changes and a mechanism to re-act to these changes to be suitable for a dynamic optimization problem.
In the literature of DCOPs detecting changes by re-evaluating the solutions is the most common change-detection approach~\cite{Nguyen20121}. The algorithm regularly re-evaluates some specific solutions (for us the first and the middle individual of the population) to detect changes in their function values or/and constraints. 
If there is a change detected then individuals of the population are re-initialized to avoid obsolete information.

Experimenting different dynamic handling mechanisms is a future topic for this work, but in this work we only focus on constraint handling mechanisms. Algorithm\ref{alg:DDE}, illustrates the pseudocode of the applied algorithm.

\begin{algorithm}[t]
\small
\begin{algorithmic}[1]
% \BEGIN
	 \STATE Create and evaluate a randomly initial population $\vec{x}_{i,G}\,\forall i, i=1, \ldots, NP$
   \FOR{$G\gets 1$ to $G_{max}$}
      \FOR{$i\gets 1$ to $NP$}
				\STATE Change detection mechanism ($\vec{x}_{i,G}$)				
				\STATE Randomly select $r0 \neq r1 \neq r2 \neq i$
				\STATE $J_{rand} = randint [1,D]$
				\FOR{$j\gets 1$ to $D$}
					\IF{$rand_j \leq CR$ Or $j = J_{rand}$}
						\STATE $u_{i,j,G} = x_{r1,j,G} + F(x_{r2,j,G} - x_{r3,j,G}) $
					\ELSE
						\STATE $u_{i,j,G} = x_{i,j,G}$
					\ENDIF
				\ENDFOR
                \STATE {Select  $u_{i,j,G}$ or $x_{i,j,G}$ based on the constraint handling}
			\ENDFOR
		\ENDFOR
%	\END
\end{algorithmic}
\caption{Dynamic differential evolution (DDE)}
\label{alg:DDE}
\vspace{.1cm}
\end{algorithm}

\section{Experimental investigations}
In this section, experimental setup will be introduced first and the results will be analyzed afterwards.
\label{sec:Experiment}
\subsection{Experimental setup}
\label{subsec:exp-setup}
The experiments are conducted for different magnitudes of hyperplane translation, changes of frequency, and a combination of hyperplane translation and rotation. The settings for $b$\_values for hyperplane translation are large: $lk$=-25, $uk$=25, medium: $lk$=-15, and $uk$=15 and small: $lk$=-5, $uk$=5 with initial value of $b$: $b\_0$=2. The settings for changes of frequency are large ($\tau=500$), medium ($\tau=1000$) and small ($\tau=2000$). The maximum number of evaluations is obtained by: $100\tau+1000$, where 100 is the number of changes in the environments and 1000 is the buffer that allows the algorithms proceed in their optimization process for the first time before a change occurs.
The first column of the tables~\ref{tab:singlecons} and~\ref{tab:statmultiple} show severities on magnitude of hyperplane translation, and the second column show severities of frequency.

The results are repeated for four artificial functions including sphere, Rastrigin, Ackley and Rosenbrock. 
Parameters of DE are chosen as $NP=20$, $CR=0.2$ and $F$ is a random number in $\in[0.2,0.8]$. The dimension of the problem is 30 for all the experiments. 

Two distinct measurements are applied for comparing the algorithms that are introduced as following. The first one is the common modified offline error~\cite{nguyen2012continuous} from the literature and the second one is our proposal for ranking the compared algorithms.

\subsubsection{Modified offline error ($\text{M\_off}\_e$)}
This measurement (Equation~\ref{eq:offlineerror}) is equal to the average of the sum of errors in each generation divided by the total number of generations~\cite{nguyen2012continuous}.

\begin{equation}
\text{M\_off}\_e= \frac{1}{G_{max}} \sum_{G = 1}^{G_{max}} e(G)
\label{eq:offlineerror}
\end{equation}
\noindent where $G_{max}$ is the number of generations computed by the algorithm and $e(G)$ denotes the error 
in the current iteration $G$ (see Equation~\ref{eq:error}):

\begin{equation}
	e(G)= |f(\vec{x}^*,t) - f(\vec{x}_{best,G},t)|
	\label{eq:error}
\end{equation}

\noindent where $f(\vec{x}^*,t)$ is the feasible best-known\footnote{This best-known is an approximation, which is the best solution found by DE for large number of evaluations at the current time.} at current time $t$, and $f(\vec{x}_{best,G},t)$ represents the best solution (feasible or infeasible) found so far at generation $G$  (for common offline error) at current time $t$. However for this modified version, in the case where the best solution is infeasible, the worst solution in the population is chosen instead of the best found. The worst solution is selected from an infeasible population as an effort to overtly encourage feasible solutions. The reason for choosing the modified offline error was because in the common offline error, constraint violation is not considered, while here our focus is to observe which one deals with the constraints more effectively.  

As this measure needs optimal solutions for each time as part of its calculation, and its not possible to have the optimum most of the time. Thus, having measures that do not need optimal solutions is appreciated in DCOPs. To this purpose, our proposed ranking procedure will be explained as follows.

\subsubsection{Ranking mechanism}
\label{subsec:RankigMech}
The following ranking procedure is build upon the feasibility rules~\cite{deb2000efficient}. Considering the best solution obtained before a change in time for algorithms $i$ and $j$, a lexicographical ordering mechanism in which the minimization of the sum of constraint violation precedes the minimization of the objective function will define the ranking of the algorithms (Equation~\ref{eq:lexicoranking}). 

\begin{equation}
%\begin{multline}
\begin{split}
\label{eq:lexicoranking}
    (f(\vec{x}_{Ai,t}), \phi(\vec{x}_{Ai,t}))  <  (f(\vec{x}_{Aj,t}), \phi(\vec{x}_{Aj,t}))
    \Leftrightarrow\\
    \begin{cases}
            f(\vec{x}_{Ai,t}) < f(\vec{x}_{Aj,t}), & \text{if } \phi(\vec{x}_{Ai,t}) = \phi(\vec{x}_{Aj,t})  \\
            \phi(\vec{x}_{Ai,t}) < \phi(\vec{x}_{Aj,t}), & \text{otherwise} \\
    \end{cases}
    \end{split}
  %  \end{multline}
\end{equation}

where $f(\vec{x}_{Ai,t})$ and $\phi(\vec{x}_{Ai,t})$, are the objective function and the sum of constraint violation (Equation~\ref{eq:sumcon}) of the best solution achieved with algorithm $i$ before a change happens respectively.
The sum of constraint violation $\phi(\vec{x},t)$ is  calculated as follows:
\begin{equation}
	\label{eq:sumcon}
	 \phi(\vec{x},t) =  \sum\limits_{i=1}^m max(0,g_i(\vec{x},t)) +  \sum\limits_{j=1}^p |h_i(\vec{x},t)|
\end{equation} 

where the inequality ($g_i(\vec{x},t)$) and equality ($h_i(\vec{x},t)$) constraints are defined in Equation~\ref{eq:searchspace}.

This lexicographical ranking procedure is applied across every test conducted for analytical testing. In each time change, the performance of the algorithms are ranked and these scores are combined into an overall performance score. Once the overall scores are calculated, the algorithms are ranked in order of the performance. We applied this ranking procedure for three algorithms, although it is adaptable to be used for comparing any number of algorithms (multi-compare).

\subsection{Experimental results}
In this section, a complementary method to compare the algorithms qualitatively is followed by the statistical test results which are divided to two cases: i) single constraint and ii) multiple constraints. 

\subsubsection{Illustration of results for sphere}
\begin{figure}[t]
        \centering          \includegraphics[width=0.5\textwidth, height=4cm]{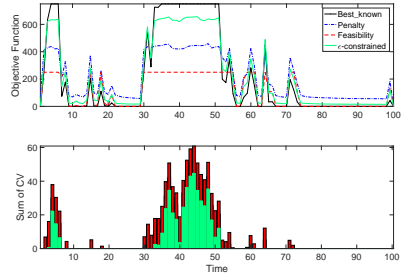}  
        \caption[]
        {\scriptsize  Sphere objective function and sum of constraint violation over-time} 
        \label{fig:funcwsumcv}
    \end{figure}

One qualitative way that helps compliment the comparison of the algorithms is to plot the objective values and sum of constraint violation for different changes in the environment, Figure~\ref{fig:funcwsumcv}. 
This figure will represent the differences between the objective function values of each algorithm (averaged across thirty runs) for the last generation before a change in time. The dot coded plot shows the best-known solution.
In the top graph, the y-axis represents the values of the objective function and in the bottom graph a bar chart is representing the sum of constraint violations for the corresponding time for each algorithm. 
This figure belongs to the relevant details of the sphere function with medium frequency and large amplitude of hyperplane translation changes. As the figure shows, the created changes by the hyperplane translation, have successfully created the new environment and put the algorithms to struggle to find the new optimum. At first look it seems feasibility and $\epsilon$-constrained are more successful to reach near-optimum solutions, however the bottom graph shows for some of the times they reach to infeasible solutions with the shown sum of constraint violation. Thus, for a better comparison of algorithms, the statistical analysis are essential.

In addition to the figure, Table~\ref{tab:singlecons} shows part of the results (25 times out of 100) of applying the benchmark for testing the sphere function with a single dynamic linear constraint. As the table shows, the hyperplane translation (medium changes of $b$\_values) changes the size of the feasible region\footnote{The feasible region is calculated by generating one million random solutions and getting the percentage of those that do not have constraint violations.} inside of the search space. In this case there is an inverse relation between the size of the feasible region for the sphere function and $b$\_value that is observable in the results. As the feasible region changes over time, new optimal points can appear, represented in the table as best\_known. The way the algorithms track these new best-known solutions is recorded across all of the runs and then the distribution is measured in the table. This allows a comparison between algorithms for specific times, although, for larger time scales individual comparisons are not preferable. The highlighted points with an asterisk shows that the relevant algorithm has sum of constraint violation other than zero for the relevant times.

\subsubsection{Single constraint}
The results of this section and the next section (multiple constraints) are based on measuring the performance of the algorithms with the ranking mechanism and the modified offline error explained in section\ref{subsec:exp-setup}; where the results are summarized in Tables~\ref{tab:statsingle} and~\ref{tab:statmultiple}. To validate the results, the 95\%-confidence Kruskal-Wallis statistical test and the Bonferroni post hoc test, as suggested in~\cite{Derrac20113} are presented. Nonparametric tests were adopted because the samples of runs did not fit to a normal distribution based on the Kolmogorov-Smirnov test.

The results of statistical tests for single constraint case presented in Table~\ref{tab:statsingle} showed in all of the cases, all the methods have significant difference with each other based on the modified offline error ($\text{M\_off}\_e$) values.  
For the hyperplane translation, it is expected that as the magnitude of changes increases from small to large, the $\text{M\_off}\_e$ values also increases as it gets harder for the algorithms to track the bigger changes. However for multi-modal functions (Rastrigin and Ackley), this trend is not observable with the exception of feasibility for Rastrigin function. The reason behind this is that with small changes, algorithms are not able to come out of their previous local optima but with larger changes this will happen resulting to smaller $\text{M\_off}\_e$ overall.
For the experiment 2, all the algorithms for all the functions showed similar trend in which $\text{M\_off}\_e$ increased as the frequency of changes increased. This is expected as with higher frequencies, algorithms have less time to reach to near optima solutions. Thus their overall deviation from the optima counted over all times increases ($\text{M\_off}\_e$).

Based on the results for $\text{M\_off}\_e$, the magnitude of the effect that hyperplane translation has on the performance of the algorithms is greater than the effect that frequency has. The magnitude in difference in $\text{M\_off}\_e$ between the respective small \& large settings was greater for hyperplane translation in every single test case. The reason behind this is that drastic changes lead to early solutions being infeasible or non-optimal, leading to larger $\text{M\_off}\_e$.

The third experiment tends to have one of the highest $\text{M\_off}\_e$, however it is usually beaten by large hyperplane translation (large setting). This is because in this experiment the hyperplane is both translating and rotating (the $b$\_value in this experiment is medium). The effect of hyperplane rotation on algorithm performance is lesser than that of translation, this leads to larger translation values affecting performance to a greater degree.

One of the limitations of $\text{M\_off}\_e$ measure is that it is biased against generations where solutions are infeasible (considers worse solution of population in case of an infeasible solution). This makes our ranking procedure better suited to dynamic environments because it only uses an algorithm's best solution for each time and considers both criteria (objective value \& sum of constraint violation) in selecting a higher performing result.

In the results, some algorithms are ranked higher than others despite having greater observed $\text{M\_off}\_e$. This discrepancy is caused by the ranking solutions selecting the single best solution for each time and then comparing the algorithms, whereas $\text{M\_off}\_e$ is measured per generation. 
Based on discrepancy of these two measures, ranking procedure will give higher priority to feasibility of the achieved solutions, while $\text{M\_off}\_e$ is more in favour of the more closer to optima solutions. So an algorithm can be chosen based of which criteria is in our priority based of these two measures.

Based on ranking results, penalty is competitive with feasibility across the functions despite usually having a higher $\text{M\_off}\_e$ due to the fact that penalty accepts more infeasible solutions (based on our adaptive penalty method this happens especially for the first generations) \& $\text{M\_off}\_e$ picks the worst solution if the best is infeasible. While, when ranking algorithms the best solution overall for each time is selected and this allows penalty to rank higher (as generations proceed penalty tend to increase penalization factor to choose feasible solutions leading to final solutions more feasible).
Achieving the third rank for $\epsilon$-constrained also is expected based on its less strict behaviour toward infeasible solutions compared to feasibility. This lead it to get more final infeasible solutions at each time, leading to its lower ranking compared to feasibility.

For Ackley function regardless of the experiment, the algorithms ranked identically relative to each other. Feasibility outperformed the others with penalty coming second and $\epsilon$-constrained coming last. For the other functions penalty and feasibility are struggling for the ranking. For the hyperplane translation and rotation, feasibility wins regardless of the tested function.
\subsubsection{Multiple constraints}
Statistical test for multiple constrained case presented in Table~\ref{tab:statmultiple} also showed all the methods have significant difference with each other. As the results in this table show multiple constraint experiments tends to have higher $\text{M\_off}\_e$ compared to the other single constraint experiments, this is due to the increased difficulty that comes with satisfying multiple constraints at the same time over a single one. 
Regardless of the experiments and severities, in the rankings feasibility is consistently as the best, the second ranking is for $\epsilon$-constrained and the last one is for penalty. This is expected as in our ranking procedure the priority is with feasibility of the solutions. In the case of multiple constraints, in most of the experiments algorithms are not able to find any feasible solution, and so based of the proposed ranking procedure they will be ranked based of lower sum of constraint violations. Thus, feasibility based on its algorithm mechanism is usually the winner in this case.

\begin{table}[t]
\centering
\caption{\scriptsize Testing benchmark for single constraint setup (sphere function)}
\label{tab:singlecons}
\scalebox{0.65}{
\begin{tabular}{l|l|c|l|l|l|l}
Time& $b$\_Values &          Feasible region(\%) & Best-Known & Penalty & Feasibility & $\epsilon$-constrained                        \\\hline

 25    &18.90    &100    &0    &68.24($\pm$8.56) &0.14($\pm$0.06)  &19.76($\pm$3.23) \\
26    &15.68    &100    &0    &65.68($\pm$8.56) &0.13($\pm$0.04)  &18.29($\pm$3.46) \\
27    &12.36    &99.99    &0    &63.22($\pm$8.30) &0.13($\pm$0.04)  &18.1554($\pm$3.35) \\
28    &14.68    &100    &0    &61.82($\pm$7.02) &0.12($\pm$0.04)  &18.0836($\pm$3.24) \\
29    &4.72    &94.90    &0    &61.73($\pm$6.98) &0.11($\pm$0.03)  &17.65($\pm$3.07) \\
30    &13.81    &100    &0    &60.67($\pm$6.92) &0.10($\pm$0.02)  &17.40($\pm$3.02)  \\
31    &3.43    &88.29    &0    &61.16($\pm$8.46) &0.10($\pm$0.02)  &17.21($\pm$2.72)  \\
32    &-3.73    &9.88    &13.86    &120.29($\pm$19.31) &50.79($\pm$3.04)  &72.16($\pm$10.62) \\
33    &3.36    &87.71    &0    &78.24($\pm$12.88) &0.23($\pm$0.10)  &23.34($\pm$5.39) \\
34    &-6.24    &1.52   &38.83    &186.09($\pm$30.30) &124.59($\pm$2.69)  &105.79($\pm$13.04) \\
35    &-2.79    &16.65    &7.79    &103.78($\pm$14.56) &31.22($\pm$2.56)  &57.34($\pm$8.82)  \\
36    &-7.59    &0.40    &57.416    &208.18($\pm$45.91) &176.86($\pm$1.20)  &129.04($\pm$12.59)  \\
37    &-17.71   &0    &312.29    &432.27($\pm$43.40)$^*$ &249.90($\pm$0.26)$^*$  &373.68($\pm$8.93)$^*$ \\
38    &-23.18    &0    &535.18    &449.60($\pm$40.75)$^*$ &249.96($\pm$0.19)$^*$  &564.60($\pm$8.40)$^*$ \\
39    &-37.95    &0    &750    &445.62($\pm$37.58)$^*$ &249.99($\pm$0.02)$^*$  &622.19($\pm$26.84)$^*$  \\
40    &-29.763    &0    &750    &446.50($\pm$39.38)$^*$ &249.95($\pm$0.17)$^*$  &631.42($\pm$27.00)$^*$  \\
41    &-16.24    &0    &262.70    &406.37($\pm$46.45)$^*$ &249.94($\pm$0.15)$^*$  &332.44($\pm$9.65)$^*$ \\
42    &-13.63    &0    &184.99    &346.09($\pm$31.43) &249.93($\pm$0.16)$^*$  &262.38($\pm$11.48)$^*$ \\
43    &-16.74    &0    &279.02   &416.34($\pm$37.78)$^*$ &249.98($\pm$0.05)$^*$  &345.43($\pm$9.60)$^*$ \\
44    &-5.81    &2.18    &33.62   &160.23($\pm$23.39) &109.16($\pm$2.41)  &98.47($\pm$11.50)  \\
45    &-9.827    &0.02    &96.12    &248.92($\pm$28.16) &249.73($\pm$1.24)$^*$  &175.17($\pm$9.74)  \\
46    &0.51    &57.04    &0    &86.80($\pm$12.70) &0.37($\pm$0.16)  &31.98($\pm$6.47)\\
47    &-1.52    &29.99    &2.29    &94.192($\pm$18.16) &11.24($\pm$2.03)  &43.25($\pm$12.46) \\
48    &-11.59    &0.00    &133.88    &282.30($\pm$33.81) &249.86($\pm$0.30)$^*$  &211.62($\pm$11.46)\\
49    &-14.86    &0    &219.82    &378.82($\pm$44.47)$^*$ &249.98($\pm$0.07)$^*$  &292.40($\pm$14.35)$^*$\\
50    &-6.50   &1.18    &42.07    &177.18($\pm$26.29) &134.64($\pm$2.25)  &111.69($\pm$10.12)
\end{tabular}
}
\end{table}

\begin{table*}[t]
\centering
\caption{\scriptsize Statistical test results for single constraint setup}
\label{tab:statsingle}
\scalebox{0.7}{
\begin{tabular}{ll|llllll}
\hline\multicolumn{8}{c}{\textbf{Function 1: Sphere}}\\\hline \multirow{2}{*}{Hyperplane translation}&\multirow{2}{*}{Frequency}&
     \multicolumn{2}{c}{Penalty}&\multicolumn{2}{c}{Feasibility}&\multicolumn{2}{c}{$\epsilon$-Constrained}\\&&
     Rank&$\text{M\_off\_e}$&Rank&$\text{M\_off\_e}$&Rank&$\text{M\_off\_e}$\\\hline
     \multirow{3}{*}{}Small&&2&123.4769($\pm$3.2231)&1&51.0917($\pm$0.41777)&3&82.0297($\pm$1.2143)\\
     Medium&1000&1&115.8218($\pm$2.7837)&2&60.0356($\pm$0.48736)&3&76.5903($\pm$1.3896)\\
     Large&&2&185.0702($\pm$3.4515)&1&158.7069($\pm$0.54108)&3&131.537($\pm$1.2637)\\\hdashline
     \multirow{2}{*}{Medium}&2000&2&99.8956($\pm$1.8684)&1&55.29($\pm$0.21682)&3&42.5212($\pm$0.55252)\\
     &500&1&130.548($\pm$2.858)&2&66.455($\pm$0.67891)&3&110.8869($\pm$2.5264)\\\hdashline
     Medium \& rotation&1000&2&144.871($\pm$2.9212)&1&104.1312($\pm$0.57572)&3&100.7042($\pm$1.4025)\\
     \hline\multicolumn{8}{c}{\textbf{Function 2: Rastrigin}}\\\hline \multirow{2}{*}{Hyperplane translation}&\multirow{2}{*}{Frequency}&
     \multicolumn{2}{c}{Penalty}&\multicolumn{2}{c}{Feasibility}&\multicolumn{2}{c}{$\epsilon$-Constrained}\\&&
     Rank&$\text{M\_off\_e}$&Rank&$\text{M\_off\_e}$&Rank&$\text{M\_off\_e}$\\\hline
     \multirow{3}{*}{}Small&&1&342.5681($\pm$5.1306)&2&74.5109($\pm$1.0088)&3&276.7856($\pm$4.3346)\\
     Medium&1000&1&302.9679($\pm$4.8389)&2&72.8356($\pm$0.88684)&3&244.0094($\pm$3.5312)\\
     Large&&2&281.6763($\pm$4.7968)&1&155.1468($\pm$0.79937)&3&221.1612($\pm$3.2412)\\\hdashline
     \multirow{2}{*}{Medium}&2000&1&274.6993($\pm$3.7236)&2&62.8506($\pm$0.40979)&3&183.5768($\pm$3.3816)\\
     &500&1&327.7439($\pm$6.1883)&2&88.3053($\pm$1.4242)&3&295.9678($\pm$4.4188)\\\hdashline
     Medium \& rotation&1000&2&295.2574($\pm$4.0704)&1&110.754($\pm$0.82186)&3&244.9176($\pm$4.2058)\\
     \hline\multicolumn{8}{c}{\textbf{Function 3: Ackley}}\\\hline \multirow{2}{*}{Hyperplane translation}&\multirow{2}{*}{Frequency}&
     \multicolumn{2}{c}{Penalty}&\multicolumn{2}{c}{Feasibility}&\multicolumn{2}{c}{$\epsilon$-Constrained}\\&&
     Rank&$\text{M\_off\_e}$&Rank&$\text{M\_off\_e}$&Rank&$\text{M\_off\_e}$\\\hline
     \multirow{3}{*}{}Small&&2&5.5978($\pm$0.073257)&1&4.6576($\pm$0.043587)&3&4.5304($\pm$0.052377)\\
     Medium&1000&2&5.6839($\pm$0.079671)&1&2.8791($\pm$0.047408)&3&4.3714($\pm$0.067268)\\
     Large&&2&4.9987($\pm$0.051107)&1&2.3524($\pm$0.051437)&3&3.697($\pm$0.055142)\\\hdashline
     \multirow{2}{*}{Medium}&2000&2&5.2359($\pm$0.081936)&1&2.1459($\pm$0.010981)&3&2.9252($\pm$0.041935)\\
     &500&2&6.1255($\pm$0.13925)&1&4.0344($\pm$0.07346)&3&5.5633($\pm$0.073377)\\\hdashline
     Medium \& rotation&1000&2&5.1843($\pm$0.053097)&1&2.9477($\pm$0.042913)&3&4.0516($\pm$0.049542)\\
     \hline\multicolumn{8}{c}{\textbf{Function 4: Rosenbrock}}\\\hline \multirow{2}{*}{Hyperplane translation}&\multirow{2}{*}{Frequency}&
     \multicolumn{2}{c}{Penalty}&\multicolumn{2}{c}{Feasibility}&\multicolumn{2}{c}{$\epsilon$-Constrained}\\&&
     Rank&$\text{M\_off\_e}$&Rank&$\text{M\_off\_e}$&Rank&$\text{M\_off\_e}$\\\hline
     \multirow{3}{*}{}Small&&2&253460.4751($\pm$9060.0677)&1&178489.771($\pm$1984.3665)&3&173310.4292($\pm$4686.6517)\\
     Medium&1000&1&247669.9881($\pm$7753.5791)&2&182819.7713($\pm$1604.4132)&3&198033.5418($\pm$2506.6281)\\
     Large&&2&617497.3469($\pm$6471.4511)&1&564491.2178($\pm$1645.71)&3&488869.177($\pm$4354.94)\\\hdashline
     \multirow{2}{*}{Medium}&2000&1&230607.7299($\pm$4989.7098)&2&184070.5758($\pm$963.1081)&3&121674.3982($\pm$1650.0393)\\
     &500&1&265362.7503($\pm$7965.0623)&2&180453.3204($\pm$1898.8888)&3&243173.5421($\pm$5108.1302)\\\hdashline
     Medium \& rotation&1000&2&403920.2601($\pm$5611.9181)&1&340830.2388($\pm$2661.1779)&3&319687.7464($\pm$3933.9349)\\
     \hline
\end{tabular}
}
\end{table*}

\begin{table*}[t]
\centering
\caption{\scriptsize Statistical test results for multiple constraint setup}
\label{tab:statmultiple}
\scalebox{0.7}{
\begin{tabular}{ll|llllll}
\hline\multicolumn{8}{c}{\textbf{Function 1: Sphere}}\\\hline \multirow{2}{*}{Hyperplane translation}&\multirow{2}{*}{Frequency}&
     \multicolumn{2}{c}{Penalty}&\multicolumn{2}{c}{Feasibility}&\multicolumn{2}{c}{$\epsilon$-Constrained}\\&&
     Rank&$\text{M\_off\_e}$&Rank&$\text{M\_off\_e}$&Rank&$\text{M\_off\_e}$\\\hline
\multirow{3}{*}{}Small&&3&185.9914($\pm$7.1305)&1&152.7258($\pm$0.68973)&2&79.158($\pm$4.383)\\
     Medium&1000&3&346.8865($\pm$3.1763)&1&388.4411($\pm$0.88649)&2&241.9969($\pm$2.6384)\\
    Large &&3&474.2232($\pm$3.7878)&1&563.3874($\pm$0.8724)&2&361.9659($\pm$1.8702)\\\hdashline
     \multirow{2}{*}{Medium}&2000&3&341.004($\pm$3.8124)&1&363.189($\pm$0.40685)&2&142.002($\pm$0.93646)\\
     &500&3&347.3142($\pm$5.2043)&1&419.4945($\pm$1.2822)&2&307.5669($\pm$2.465)\\\hdashline
     Medium+ rotation&1000&3&437.8225($\pm$3.8624)&1&511.3112($\pm$0.70724)&2&337.0053($\pm$1.9905)\\
     \hline\multicolumn{8}{c}{\textbf{Function 2: Rastrigin}}\\\hline \multirow{2}{*}{Hyperplane translation}&\multirow{2}{*}{Frequency}&
     \multicolumn{2}{c}{Penalty}&\multicolumn{2}{c}{Feasibility}&\multicolumn{2}{c}{$\epsilon$-Constrained}\\&&
     Rank&$\text{M\_off\_e}$&Rank&$\text{M\_off\_e}$&Rank&$\text{M\_off\_e}$\\\hline
     \multirow{3}{*}{}Small&&3&276.756($\pm$9.0855)&1&136.6664($\pm$0.63241)&2&212.4283($\pm$2.4608)\\
     Medium&1000&3&231.7457($\pm$4.6501)&1&351.4985($\pm$0.84996)&2&171.6611($\pm$2.4179)\\
     Large&&3&201.1431($\pm$3.8658)&1&504.2874($\pm$0.82039)&2&140.2549($\pm$1.6473)\\\hdashline
     \multirow{2}{*}{Medium}&2000&3&222.7672($\pm$4.9358)&1&344.7768($\pm$0.39287)&2&121.304($\pm$2.1041)\\
     &500&3&237.9647($\pm$4.7002)&1&371.1831($\pm$1.4997)&2&210.8059($\pm$3.0099)\\\hdashline
     Medium\& rotation&1000&3&218.7773($\pm$3.4863)&1&460.518($\pm$1.0985)&2&156.1407($\pm$2.1522)\\
     \hline\multicolumn{8}{c}{\textbf{Function 3: Ackley}}\\\hline \multirow{2}{*}{Hyperplane translation}&\multirow{2}{*}{Frequency}&
     \multicolumn{2}{c}{Penalty}&\multicolumn{2}{c}{Feasibility}&\multicolumn{2}{c}{$\epsilon$-Constrained}\\&&
     Rank&$\text{M\_off\_e}$&Rank&$\text{M\_off\_e}$&Rank&$\text{M\_off\_e}$\\\hline
     \multirow{3}{*}{}Small&&3&2.932($\pm$0.084437)&1&3.3296($\pm$0.013323)&2&2.1951($\pm$0.030209)\\
     Medium&1000&3&3.1001($\pm$0.050789)&1&1.9236($\pm$0.032673)&2&2.0158($\pm$0.02762)\\
     Large&&3&2.1871($\pm$0.027546)&1&0.91659($\pm$0.018204)&2&1.2362($\pm$0.016717)\\\hdashline
     \multirow{2}{*}{Medium}&2000&3&2.9026($\pm$0.055156)&1&1.5584($\pm$0.0096997)&2&1.3741($\pm$0.022489)\\
     &500&3&3.2168($\pm$0.043058)&1&2.3757($\pm$0.04413)&2&2.6736($\pm$0.0386)\\\hdashline
     Medium \& rotation&1000&3&2.4982($\pm$0.047531)&1&1.3341($\pm$0.015384)&2&1.5349($\pm$0.024865)\\
     \hline\multicolumn{8}{c}{\textbf{Function 4: Rosenbrock}}\\\hline 
     \multirow{2}{*}{Hyperplane translation}&\multirow{2}{*}{Frequency}&
     \multicolumn{2}{c}{Penalty}&\multicolumn{2}{c}{Feasibility}&\multicolumn{2}{c}{$\epsilon$-Constrained}\\&&
     Rank&$\text{M\_off\_e}$&Rank&$\text{M\_off\_e}$&Rank&$\text{M\_off\_e}$\\\hline
     \multirow{3}{*}{}Small&&3&671088.1711($\pm$30354.7562)&1&455385.3979($\pm$3183.3957)&2&334609.919($\pm$15684.2807)\\
     Medium&1000&3&1422991.1821($\pm$8921.8589)&1&1379550.3729($\pm$2485.4699)&2&1050610.912($\pm$11110.1984)\\
     Large&&3&2098319.2063($\pm$8777.8434)&1&2074607.4329($\pm$3342.8739)&2&1647633.1671($\pm$7008.7808)\\\hdashline
     \multirow{2}{*}{Medium}&2000&3&1411721.3622($\pm$14660.1552)&1&1280034.8117($\pm$979.2411)&2&646148.5487($\pm$5147.4104)\\
     &500&3&1406504.5974($\pm$14776.6532)&1&1494601.7865($\pm$3355.8843)&2&1280330.1804($\pm$5866.8826)\\\hdashline
     Medium\& rotation&1000&3&1909408.8837($\pm$11346.388)&1&1895323.5581($\pm$3172.1567)&2&1519570.3026($\pm$7386.3231)\\
     \hline   
\end{tabular}
}
\end{table*}

\section{Conclusion and future work} \label{sec:conclusion}
In this paper a framework has been proposed to generate benchmarks for testing algorithms in DCOPs. Our proposed framework can produce multiple benchmarks to be applied for testing any function and for any number of changes and dimension in the optimization problem. The changes in the environment are imposed by translation and rotation of the hyperplane in single and multiple linear constraints.
For testing our benchmark, three constraint handling techniques have been applied and compared (penalty, feasibility and $\epsilon$-constrained) with differential evolution algorithm. A procedure for ranking the algorithms that is based on the feasibility rules was proposed to analyze the results and compare the algorithms behaviour.
Implementing different functions showed that our proposed benchmark can be applied to test any function in DCOPs effectively.
Moreover, the results showed that created changes had an observable effect on the performance of the compared algorithms. 
For future work the proposed benchmark would need to be used with more advanced algorithms as the current constraint handling techniques struggle with dynamism. 
\section*{Acknowledgment}
This work has been supported through Australian Research Council (ARC) grant DP160102401.

%\bibliographystyle{IEEEtran}
%\bibliography{mybibfile}
  %abbrv
	\bibliographystyle{IEEEtran}
\bibliography{main}

\end{document}